\documentclass[a4paper,twoside]{article}

\usepackage{epsfig}
\usepackage{graphicx}
\usepackage{subcaption}
\usepackage{calc}
\usepackage{amssymb}
\usepackage{amstext}
\usepackage{amsmath}
\usepackage{amsthm}
\usepackage{multicol}
\usepackage{pslatex}
\usepackage{apalike}
\usepackage[bottom]{footmisc}
\usepackage{hyperref}
\usepackage{SCITEPRESS}     

\begin{document}

\title{Unifying Human Motion Synthesis and Style Transfer with Denoising Diffusion Probabilistic Models}

\author{\authorname{Ziyi Chang\orcidAuthor{0000-0003-0746-6826}, Edmund J. C. Findlay, Haozheng Zhang and Hubert P. H. Shum\orcidAuthor{0000-0001-5651-6039}\thanks{Corresponding author}}
\affiliation{Department of Computer Science, Durham University, Durham, UK}
\email{\{ziyi.chang, edmund.findlay, haozheng.zhang, hubert.shum\}@durham.ac.uk}
}

\keywords{Diffusion Model, Animation Synthesis, Human Motion}

\abstract{Generating realistic motions for digital humans is a core but challenging part of computer animations and games, as human motions are both diverse in content and rich in styles. While the latest deep learning approaches have made significant advancements in this domain, they mostly consider motion synthesis and style manipulation as two separate problems. This is mainly due to the challenge of learning both motion contents that account for the inter-class behaviour and styles that account for the intra-class behaviour effectively in a common representation. To tackle this challenge, we propose a denoising diffusion probabilistic model solution for styled motion synthesis. As diffusion models have a high capacity brought by the injection of stochasticity, we can represent both inter-class motion content and intra-class style behaviour in the same latent. This results in an integrated, end-to-end trained pipeline that facilitates the generation of optimal motion and exploration of content-style coupled latent space. To achieve high-quality results, we design a multi-task architecture of diffusion model that strategically generates aspects of human motions for local guidance. We also design adversarial and physical regulations for global guidance. We demonstrate superior performance with quantitative and qualitative results and validate the effectiveness of our multi-task architecture.
}

\onecolumn \maketitle \normalsize \setcounter{footnote}{0} \vfill

\section{\uppercase{Introduction}}
\label{sec:introduction}

The generation of realistic human motions has been an important but challenging task in computer graphics and vision. Compared to motion capture, generating motions allows obtaining data with affordable cost and abundant amount. Furthermore, the generated motions have a wide range of potential applications, e.g. making animations and games. As human motions are inherently diverse in both contents and styles, generating realistic motions is also challenging. For example, realistic human motions have a variety of contents, such as walking and running. Even within each content, motions vary from each other, e.g. strutting walking and depressed walking.

\begin{figure}[!h]
  \centering
   {\epsfig{file = 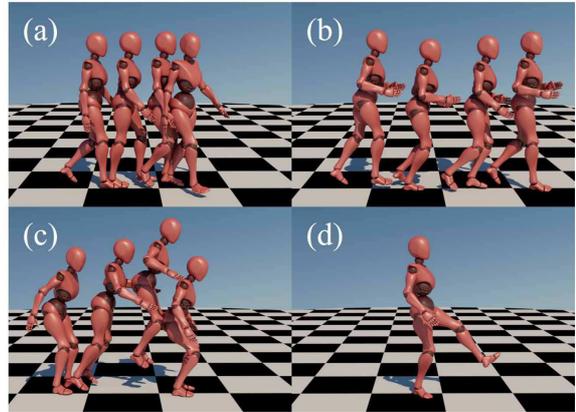, width = 1.0\linewidth}}
  \caption{Sampled results. (a) is walking. (b) is running. (c) is proud jumping. (d) is angry kicking.}
  \label{fig:styled_samples}
 \end{figure}

While recent years have witnessed some significant advances made by many deep learning approaches towards this domain, these approaches mainly consider motion synthesis and style manipulation as two separate tasks. Approaches to motion synthesis are usually concentrated on the generation of various contents \cite{holden2017phase,mourot2022survey}, while the style transferring approaches focus on the manipulation of motion styles \cite{aberman2020unpaired,ye2021human}. Although methods from the two separate tasks can be sequentially applied, the generated motions are likely to be better if the common space is modelled jointly. 

The main challenge of generating realistic human motions is to effectively model the inter-class (i.e. content) and intra-class (i.e. style) behaviours in a common representation. Human motions are composed of contents and styles \cite{aberman2020unpaired}. The difference in contents dominates different inter-class behaviours, while the variation of styles accounts for the various intra-class motion behaviours. Modelling inter-class and intra-class behaviours of realistic human motions requires neural networks to learn a common representation in the latent space. The underlying distribution of such a latent space is greatly wider than that of modelling only one component, as both content and styles are inherently diverse. The increased diversity of data requires a higher capacity in neural networks to model the potential probabilistic distributions. However, traditional methods can only focus on learning the distribution of one component due to the limited capacity of their generative models \cite{bond2021deep} as they suffer from sequential generation~\cite{martinez2017human}, mode collapse~\cite{dong2020adult2child}, prior distribution assumption~\cite{ling2020character} or specially-designed architecture~\cite{henter2020moglow}.

We propose a diffusion-based solution to unify human motion synthesis and style transfer. Because its mode coverage ability is wider than other previous neural networks, it is beneficial to model content and style in a common representation. Specifically, denoising diffusion probabilistic model (DDPM) is a diffusion-based generative model with high capacity due to the injection of stochasticity in the learning process, which is inspired by non-equilibrium thermodynamics in physics. To achieve high-quality motion synthesis and style-transfer results, we design a multi-task DDPM architecture that strategically models different aspects of realistic human motions, including joint angles, global movement trajectories, supporting foot patterns and physical regulations. Compared to the original DDPM \cite{ho2020denoising}, the multi-task architecture in our design increases the ability to model the data structure of human motions. Apart from only predicting the noise, our multi-task architecture further predicts other aspects of human motions with additional neural networks. To enhance the synthesized motions to be realistic globally, we also leverage adversarial training to coordinate different predictions from multiple tasks and ensure them to be harmonious with each other. In addition to adversarial regulation, we also leverage physical regulations to achieve global coherent motions.

We demonstrate the superior performance of our proposed solution by both quantitative and qualitative experiments on the dataset proposed by \cite{xia2015realtime} and validate the effectiveness of our design by ablation study. In addition, the Fréchet inception distance (FID) \cite{heusel2017gans} has been reported to measure the distribution difference between the original and generated motion datasets. 
Our method achieves the lowest FID, signifying superior performance. We also provide several synthesized results by figures to show the quality of our method. We also validate different components of our multi-task architecture by ablation study. 

An existing work presents preliminary results of synthesizing human motions by diffusion models \cite{findlay2022denoising} where a barebone prototype to show the potential feasibility that walking motion with different styles can be synthesised with diffusion models. However, \cite{findlay2022denoising} suffers from low motion quality in non-walking motion due to the simple network design. In this paper, we present an improved solution that is capable of generating high-quality motion of different contents with styles. Furthermore, we showcase the improvement over \cite{findlay2022denoising} in our experiments. This paper presents the following contributions:
\begin{itemize}
    \item We present a single-stage pipeline unifying human motion synthesis and style transfer for high-quality motion creation. The source code is open on \url{https://github.com/mrzzy2021/StyledMotionSynthesis}
    \item To effectively represent the coupled representation of both inter-class motion contents and intra-class motion styles in a common latent space, we propose a denoising diffusion probabilistic model solution that has a large learning capacity for modelling the diverse data structure.
    \item To generate high-quality results, we propose a multi-task network architecture that leverages both local guidance, including joint angles, movement trajectories and supporting foot patterns, and global guidance, including physical and adversarial regulations.
\end{itemize}

\section{\uppercase{Related Work}}

Our work is mainly related to two research fields, neural motion generation and generative models. First, we will review the recent advances in neural motion generation, mainly including human motion synthesis and motion style transfer. In the generative models part, we will mainly discuss diffusion models.

\subsection{Neural Motion Generation}

Most motion synthesis works are prediction-based. Given a past pose or partial sequence, a model will predict future poses. Recent advances in deep generative models such as Generative Adversarial Networks (GANs), Variational Autoencoders (VAEs), long short-term memory (LSTM) models and Flow-based models have seen a strong performance with this approach \cite{mourot2022survey}. 

MoGlow \cite{henter2020moglow} used an LSTM-based normalising flow model inspired by recent applications of such models to video predicting. When provided with a context of previous poses and a control signal containing the character's future angular velocity, the model can generate the next pose in the sequence. Characters can accurately follow various trajectories for long and short-term motions by using this approach. However, this method can only generate the motions one frame at a time, which is not ideal for real-time applications such as computer graphics. The generated motions are diverse but often suffer inaccuracies, such as foot sliding and joint crossover. Wen et al. also show that this model architecture performs strongly for unsupervised motion style transfer. Although the results of flow-based models are promising, they are often parameter inefficient compared to other generative models such as GANs.

Improving on the control signals from MoGlow \cite{henter2020moglow} for human locomotion, Ling et al. propose Motion VAEs \cite{ling2020character}. Using a two-stage approach, the authors first train an encoder and mixture-of-experts decoder on the pose sequences dataset of locomotion data. They then train a reinforcement learning algorithm to predict the latent space input to be passed to the decoder based on the previous pose and a control signal. The final method is able to control a character using a joystick as well as by providing a trajectory or a target. Adding noise during the second stage training process allows for diversity when generating controlled inputs. However, the algorithm demonstrates bias since the generated data tend to have right-side behaviour whilst the training data contains equal amounts of right and left-handed data.

The action-conditioned motion generation is typically considered a more difficult task than motion prediction-based synthesis since less input information is supplied to the model. The Text2Action \cite{ahn2018text2action} uses a recurrent neural network (RNN) encoder to take a sentence as input and output an embedding. This embedding is then used to condition a generator RNN which will output a motion based on the original sentence. A separate recurrent discriminator is also trained to minimise perceptual differences between generated motion and the training data. The system struggled to achieve sufficient quality motions when trained end-to-end and relied on a multistage training procedure. ACTOR \cite{petrovich2021action} builds on Text2Action by introducing a transformer-based VAE approach. Different from \cite{ahn2018text2action}, this method does not require multistage training and the training is more stable as it is not adversarial. The transformer will generate the appropriate pose sequence when supplied with the desired action. This model significantly outperforms their baseline gated recurrent unit (GRU) model on multiple datasets. Due to the autoregressive nature of the decoder, the method can be adapted to generate actions of varying lengths.

\subsection{Diffusion Models}

Diffusion models are firstly proposed by \cite{sohl2015deep} for data modelling. The diffusion phenomenon is driven by a stochastic dynamic system in physics and destroys the original data structure. Aiming to model the diffusion and its reverse process, \cite{sohl2015deep} proposes diffusion models and uses a neural network to recover the data distribution. \cite{sohl2015deep} uses the variational lower bound to optimize the negative log-likelihood of diffusion models.

The denoising diffusion probabilistic model (DDPM) is firstly proposed by \cite{ho2020denoising}. Instead of using variational lower bound, \cite{ho2020denoising} simplifies the loss function by predicting the added noise. As reported by \cite{ho2020denoising}, predicting the added noise achieves better results than predicting the intermediate data. Many improvements have also been proposed to DDPM. \cite{nichol2021improved} proposes to predict the covariance matrix during the reverse process and implements a cosine noise schedule instead of a linear schedule. \cite{dhariwal2021diffusion} proposes a guidance-based diffusion model where auxiliary classifiers have been pre-trained to provide guidance to the generation. \cite{song2020denoising} proposes a denoising diffusion implicit model (DDIM) to accelerate the sampling speed of DDPM. \cite{kawar2022denoising} proposes a denoising diffusion restoration model (DDRM) to specialize DDPM in image restoration problems.

Hybrid approaches which combine a DDPM with another type of generative model (e.g., VAEs or GANs) have shown early promise in addressing the issues with diffusion models. \cite{vahdat2021score} propose to train a VAE and then train a diffusion model in the latent space. This method surpasses all previous diffusion models when trained on image data and only requires around 100 sampling steps due to the reduced complexity of the latent space compared to the original data. However, training both models is a multistage process as the VAE needs to be trained separately and then trained with the diffusion model. It results in more computational resources than a standard diffusion model to achieve these results. Denoising diffusion GANs \cite{xiao2021tackling} use a reformulation of the diffusion process where the original data is estimated instead of the noise added. As a result, the model is less prone to mode collapse than GAN models and is also more stable to train. However, the quality of the generated samples is still slightly lower than those generated by previous diffusion models.

Most recently, diffusion models have achieved impressive results for text-to-image generation. GLIDE \cite{nichol2021glide} uses an encoding transformer to input a caption to a diffusion model. Imagen \cite{saharia2022photorealistic} proposes to leverage generic large language models that are pretrained on text-only corpora for text-to-image translation. DALLE-2 \cite{ramesh2022hierarchical} proposes a two-stage model for text-guided image generation. Their two stages include a prior that generates a CLIP image embedding given a text caption, and a decoder that generates an image conditioned on the image embedding. Such models for text-to-image transition usually are large. For example, \cite{nichol2021glide} has 3.5 billion parameters in size and is extremely computationally intensive to train. However, when combined with another 1.5 billion parameters, the upsampling diffusion model can generate high-resolution samples when provided with a text input.

Diffusion models have excelled at a range of tasks on many different data domains. They offer many benefits over other model classes, providing both high-quality and highly diverse samples at the cost of sampling time. However, despite their success, they have not yet been applied to motion data.

\section{\uppercase{Methodology}}

Generating realistic human motions is challenging due to the diversity of the common representation for the coupled content-style information. Motions are composed of inter-class contents and intra-class styles \cite{aberman2020unpaired}. Previous methods do not have sufficient learning capacity for modelling the coupled representation in a common latent space, such that they learn content and style modelling separately as motion synthesis and style transfer. However, separating as two tasks usually results in sub-optimal motions and under-explored joint distribution space.

\begin{figure*}[!h]
  \centering
   {\epsfig{file = 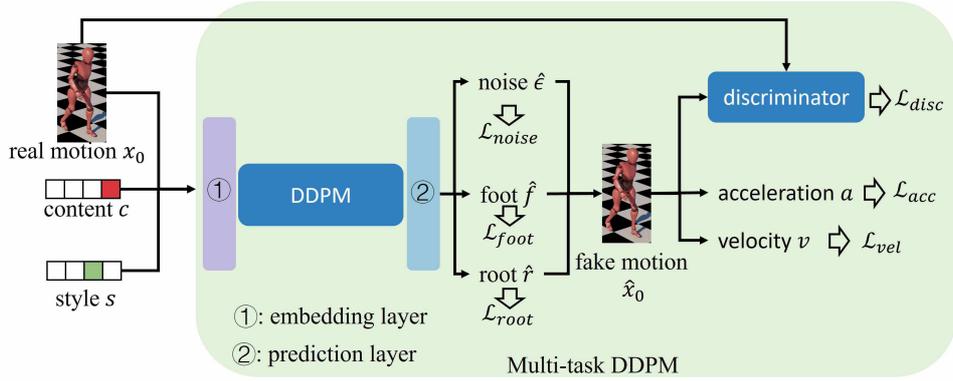, width = 0.8\linewidth}}
  \caption{An overview of our proposed framework.}
  \label{fig:pipeline}
 \end{figure*}

To tackle the challenge, we propose our denoising diffusion probabilistic model (DDPM) solution for styled motion synthesis as shown in Figure \ref{fig:pipeline}. DDPM is a diffusion-based generative model that injects and removes Gaussian noise progressively. The introduction of stochasticity in the learning process brings a larger capacity to the neural network. Compared to previous methods, DDPM is a more suitable candidate for styled motion synthesis as it does not suffer from the lack of generation diversity. Our proposed DDPM solution facilities an integrated, end-to-end framework with increased learning capacity. The motion creation benefits more from our single-stage pipeline when compared to previously two-stage solutions. The integration enables direct optimization over the joint common representation space, and the underlying manifold is better explored to obtain better performance.

The training process of our proposed pipeline shown in Figure \ref{fig:pipeline} is designed to be a multi-task architecture. The pipeline is optimized over a set of predictions on different motion aspects, including joint angles, foot patterns, global movements and physical regulations. Additionally, a discriminator is applied for adversarial optimization over our pipeline.

We unify human motion synthesis and style transfer into our proposed end-to-end pipeline. The framework leverages DDPM to increase the learning capacity to model the diversity in motion data. The architecture of our proposed pipeline is designed to be multi-task with a set of corresponding auxiliary losses for a high-quality generation.

\subsection{Problem Formulation}

Generating realistic human motions is an important but challenging task. Modelling human motions requires the neural network to learn the inter-class motion contents and the intra-class motion styles simultaneously. The joint distribution of contents and styles describes the underlying motion manifold in the latent space. However, the great diversity of both human contents and styles demands a high learning capacity of neural networks to maintain generation diversity, leading to a separate treatment in previous studies.

\begin{figure}[!h]
  \centering
   {\epsfig{file = 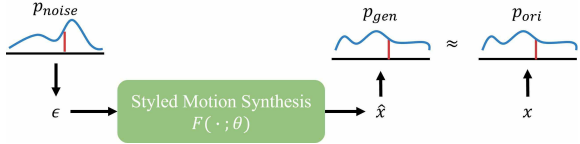, width = 1.0\linewidth}}
  \caption{Our proposed end-to-end framework for the styled motion synthesis task.}
  \label{fig:framework}
 \end{figure}

We propose a new problem called styled motion synthesis from an integration view and a framework, as shown in Figure \ref{fig:framework}. Previous studies have been conducted in developing neural networks for human motion synthesis or motion transfer, separately. The lack of unification formulation brings incoordination across neural networks in practice, undermining the quality of generated results. Instead of optimizing separate neural networks for human motion synthesis and motion style transfer, the new problem concerns the integration of the two tasks. Styled motion synthesis demands a neural network to generate motions with styles from noise distribution that is easy to sample from in an end-to-end manner.

To formally propose this problem, we provide a formulation. Let $x$ and $\hat{x}$ be our original motion data of all frames and generated motion data of all frames, respectively. The distributions of the original motion and generated motion are denoted as $p_{ori}$ and $p_{gen}$. The neural network $F(\cdot)$ in styled motion synthesis starts from a noise $\epsilon$ that is sampled from a distribution $p_{noise}$ and generates $\hat{x}$ that follows the distribution of original data $x$. Styled motion synthesis is formulated as follows:
\begin{equation}
    \hat{x}=F(\epsilon;\theta)\sim p_{gen}\approx p_{ori},
\end{equation}
where the noise is sampled from the noise distribution $\epsilon\sim p_{noise}$ and $\theta$ denotes the parameters to be optimized in the neural network $F(\cdot)$.

\subsection{DDPM Backbone}

We propose to use the denoising diffusion probabilistic model since it has the potential to possess enough capacity to model diversity compared with existing works. Realistic human motions are inherently diverse as motions differ from each other in terms of the inter-class contents and the intra-class styles. Modelling motions requires a larger capacity of networks in data modelling. Previous methods such as \cite{aberman2020unpaired} only focus on modelling one component as their networks are limited in capacity. This limitation usually leads to a lack of diversity in their generated results. 

Compared with the existing methods, the denoising diffusion probabilistic model (DDPM) \cite{sohl2015deep,ho2020denoising} has an increased learning capacity due to the stochasticity introduced in the neural network. DDPM is essentially designed to formulate the generation process as a stochastic system. Both the injection and noise removal are designed to be tractable for changing probabilistic distributions. In addition, the stochastic characteristic in DDPM extends the exploration scope of the latent space, facilitating the network to model data diversity.

The training process of DDPM is designed to gradually inject (the first line in Eq. \ref{eq:ddpm}) and remove (the second line in Eq. \ref{eq:ddpm}) a Gaussian noise $\epsilon\sim\mathcal{N}(0,I)$ to the input data $x_0=x$ within $t$ diffusion steps. The optimization (see the third line in Eq. \ref{eq:ddpm}) is based on minimizing the prediction of what noise has been added. The whole training process is formulated as follows:
\begin{equation}
    \begin{cases}
        p_t(x_t|x_0)=\mathcal{N}(x_t;\sqrt{\bar{\sigma}_t}x_0,(1-\bar{\sigma}_t)I),\\
        p_t(x_{t-1}|x_t)=\mathcal{N}(x_{t-1};\frac{1}{\sqrt{1-\sigma_t}}(x_t-\frac{\sigma_t}{\sqrt{1-\bar{\sigma}_t}}\hat{\epsilon}),\sigma_tI),\\
        \mathcal{L}=\mathop{\mathbb{E}}_{t, x_0,\epsilon}||\epsilon - \hat{\epsilon}||_2^2,
    \end{cases}
    \label{eq:ddpm}
\end{equation}
where $\sigma_t\in(0,1)$ is the noise schedule, $\bar{\sigma}_t=\prod_{i=1}^t1-\sigma_i$, $\hat{\epsilon}$ is the prediction of the neural network and $\mathcal{L}$ is the loss function of DDPM in \cite{ho2020denoising}. When training is completed, the generation process starts from an isotropic Gaussian noise $x_T\sim\mathcal{N}(0,I)$ and iteratively compute the previous state $x_{t-1}$ over $T$ diffusion steps, which is formulated as:
\begin{equation}
    x_{t-1}=\frac{1}{\sqrt{1-\sigma_t}}(x_t-\frac{\sigma_t}{\sqrt{1-\bar{\sigma_t}}}\hat{\epsilon})+\sigma z,
\end{equation}
where $z\sim\mathcal{N}(0,I)$ is a Gaussian noise.

\subsection{Multi-task DDPM for Styled Motion Synthesis}

\subsubsection{Local Guidance}
To achieve high-quality generated results, we propose to leverage a multi-task DDPM architecture for styled human motion synthesis. Directly applying DDPM to motion synthesis does not produce high-quality results due to the lack of consideration of other inherent aspects of motions, such as foot patterns. 

Our proposed multi-task design considers the optimization of trajectory movements, foot patterns and joint angle predictions, while the original DDPM only minimizes the difference between the predicted noise and the ground truth noise. We provide DDPM with extra guidance from other aspects of motion by the proposed multi-task design. With such guidance, the learnt distribution of coupled contents and styles is close to the manifold of realistic motions in the common latent space.

\begin{figure}[!h]
  \centering
   {\epsfig{file = 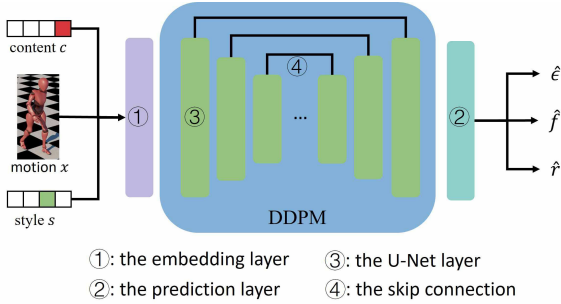, width = 1.0\linewidth}}
  \caption{Our proposed multi-task DDPM pipeline for styled motion synthesis.}
  \label{fig:multitask}
 \end{figure}

As shown in Figure \ref{fig:multitask}, our multi-task DDPM takes a motion clip $x_0=x$, the content $c$ and the style $s$ as inputs. We represent the motion $x$ with joint angles. Both the content $c$ and the style $s$ are converted into one-hot embeddings with embedding layers before they are used in DDPM. We apply the U-Net with attention and skip connections in our DDPM to model the manifold conformed by motions $x$, the inter-class contents $c$ and the intra-class styles $s$.

Our multi-task DDPM produces estimation on the noise $\hat{\epsilon}$, global movements $\hat{r}$ and supporting foot patterns $\hat{f}$ based on the three inputs. For supporting foot patterns, we use binary values to indicate a foot contacts to the floor. We leverage the noise prediction $\hat{\epsilon}$ instead of directly predicting latent joint angles $\hat{x}$ for joint angles because DDPM shows better performance by predicting noise $\hat{\epsilon}$ than latent intermediate result $\hat{x}$ \cite{ho2020denoising}. To optimize the three predictions, we formulate loss functions as:
\begin{align}
    \mathcal{L}_{noise}&=\mathop{\mathbb{E}}_{x_0, t, s, c}||\epsilon - \hat{\epsilon}||_2^2,\\
    \mathcal{L}_{foot}&=\mathop{\mathbb{E}}_{x_0, t, s, c}||f - \hat{f}||_2^2,\\
    \mathcal{L}_{root}&=\mathop{\mathbb{E}}_{x_0, t, s, c}||r - \hat{r}||_2^2.
\end{align}

\subsubsection{Global Guidance}

Apart from separately providing guidance on different aspects, we also propose to apply global guidance for the conformity of all previous aspects. We propose to apply physical regulations and a discriminator based on a reconstruction formulation in DDPM to enhance the generated motions to be realistic.

\begin{figure}[!h]
  \centering
   {\epsfig{file = 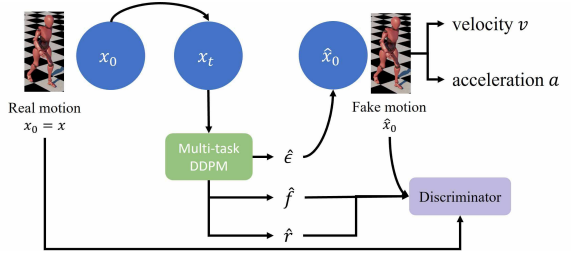, width = 1.0\linewidth}}
  \caption{Our proposed multi-task conditional DDPM pipeline for styled motion synthesis.}
  \label{fig:blended}
 \end{figure}

For our global guidance, we propose to utilize a reconstruction formulation (Eq. \ref{eq:recon}), which is derived from the arbitrary query property (Eq. \ref{eq:forwardexplicit}) in DDPM. The process of gradually adding Gaussian noise to the input $x_0$ as described in the first line of Eq. \ref{eq:ddpm} allows the arbitrary query on intermediate state $x_t$ at any diffusion step $t$, which is written as:
\begin{equation}\label{eq:forwardexplicit}
    x_t=\sqrt{\bar{\sigma}_t}x_0+\sqrt{1-\bar{\sigma}_t}\epsilon_t.
\end{equation}
The arbitrary query property is denoted by the arrow from $x_0$ to $x_t$ in Figure \ref{fig:blended}.
Our neural network makes predictions on the added noise $\epsilon_t$, which enables us to estimate $x_0$ reversely. Therefore, based on the prediction $\hat{\epsilon}$, the reconstruction estimation $\hat{x}_0$ is written as:
\begin{equation}\label{eq:recon}
    \hat{x}_0=\frac{x_t-\sqrt{1-\bar{\sigma}_t}\hat{\epsilon}}{\sqrt{\bar{\sigma}_t}},
\end{equation}
which is denoted as the arrow from $\hat{\epsilon}_t$ to $\hat{x}_0$ in Figure \ref{fig:blended}. By the reconstruction formulation, we inversely obtain an estimation of motions $\hat{x}_0$ at any diffusion step $t$.

We propose to fulfil the physical regulations based on the estimated motion $\hat{x}$. These physical regulations enforce the physical properties in real world, encouraging natural and coherent motions to be generated by our network. Specifically, the state of joints in two consecutive frames is approximately the same as humans move consecutively in real world. Furthermore, the velocity of joints between two consecutive frames in realistic human motions does not change abruptly in real world following the inertia constraint. Based on the two observations, we propose to regulate the velocity and acceleration in our estimated motion $\hat{x}$ to encourage natural and coherent results. We use the rotation of joints, denoted as $j_t$, to calculate the velocity and acceleration. Therefore, we propose the following two loss functions:
\begin{align}
    \mathcal{L}_{vel}&=\mathop{\mathbb{E}}_{t,j\in\hat{x}}||j_t - j_{t-1}||_2^2,\\
    \mathcal{L}_{acc}&=\mathop{\mathbb{E}}_{t,j\in\hat{x}}||j_{t}-2j_{t-1}+j_{t-2}||_2^2.
\end{align}

Apart from physical regulations, we also propose an adversarial design to enhance the consonance between these tasks. We design a discriminator to provide the global harmony guidance. As we predict the joint angles $\hat{x}$, global movements $\hat{r}$ and supporting foot patterns $\hat{f}$ separately in local guidance, the proposed discriminator examines whether the three estimated values are harmoniously combined into a natural motion clip. Following the adversarial training paradigm, our discriminator is trained to distinguish the generated motion clip and the ground truth motion clip. The adversarial loss is written as:
\begin{equation}
    \mathcal{L}_{disc} = ||D(x_0,r,f) - 1||^2_2 + ||D(\hat{x}_0,\hat{r},\hat{f}) - 0||_2^2.
\end{equation}
Overall, our multi-task DDPM is optimized by a set of respective loss functions on multiple tasks. The overall loss is written as:
\begin{align}
    \mathcal{L}=&\lambda_1\mathcal{L}_{noise}+\lambda_2\mathcal{L}_{foot}+\lambda_3\mathcal{L}_{root}\nonumber\\
    &+\lambda_4\mathcal{L}_{disc}+\lambda_5\mathcal{L}_{vel}+\lambda_6\mathcal{L}_{acc},
\end{align}
where $\lambda_1=\lambda_2=\lambda_3=\lambda_4=1$ and $\lambda_5=\lambda_6=0.01$.

\section{\uppercase{Experimental Results}}
We train our method on a public motion dataset~\cite{xia2015realtime} and evaluate the performance quantitatively and qualitatively. We compared our method with the initial DDPM method \cite{ho2020denoising} and the existing DDPM-based method \cite{findlay2022denoising} for evaluating performance quantitatively. In addition, we provide several generated motions for qualitative evaluation on generating motions with different contents and styles. To validate the effectiveness of each component in our proposed multi-task architecture design, we conduct a comprehensive ablation study.

\subsection{Implementation Details}
We trained all models on an NVIDIA RTX 3090 GPU and used 32-bit floating-point arithmetic. The training process can be completed within approximately one day. The generation process for each instance takes around 20 seconds. The hyperparameters in our experiments are shown in Table \ref{tab:hyperparams}.
\begin{table}[htbp]
\renewcommand{\arraystretch}{1.3}
\caption{Hyperparameters}
\label{tab:hyperparams}
\centering
\begin{tabular}{cc}
\hline
Learning Rate & 0.0002\\
\hline
Discriminator Learning Rate & 0.0001\\
\hline
Adam $\beta_1$ & 0.9\\
\hline
Adam $\beta_2$ & 0.999\\
\hline
Adam $\epsilon$ & $1e^{-8}$\\
\hline
Batch Size & 128\\
\hline
number of timesteps $T$ & 1000\\
\hline
EMA decay rate $m$ & 0.9999\\
\hline
\end{tabular}
\end{table}

We train our model with a publicly available dataset proposed by \cite{xia2015realtime}. The dataset contains six content classes and eight different styles for each type of motion. This dataset offers a wide range of styles and contents, allowing us to assess our model's diversity by combining different styles and contents during motion generation. Additionally, we apply a Gaussian filter and inverse kinematics as the post-processing.

\subsection{Quantitative Comparison}
We choose to evaluate our models using the Fréchet inception distance (FID) \cite{heusel2017gans}. This metric compares the distribution of motions generated by our diffusion model with the distribution of motions in our dataset. Given a set of generated motions from our model and a set of real motions from our dataset, we calculate a multidimensional Gaussian distribution from the features of a neural network for these datasets, which are represented as $\mathcal{N}(\mu_g,\Sigma_g)$ for our generated data and $\mathcal{N}(\mu_r,\Sigma_r)$ for our real data. We use the penultimate features from our pre-trained classifier for content motions. The FID metric is calculated by:
\begin{equation}
    FID = ||\mu_r - \mu_g||^2_2 + tr(\Sigma_g + \Sigma_r - 2(\Sigma_r\Sigma_g)^{\frac{1}{2}}).
    \label{FID}
\end{equation}

\begin{figure}[!h]
  \centering
   {\epsfig{file = 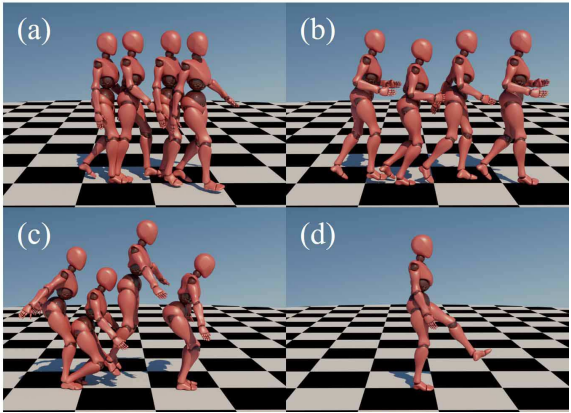, width = 1.0\linewidth}}
  \caption{Generated motions with different contents. (a) is walking. (b) is running. (c) is jumping. (d) is kicking.}
  \label{fig:content}
 \end{figure}

\begin{figure*}[!h]
  \centering
   {\epsfig{file = 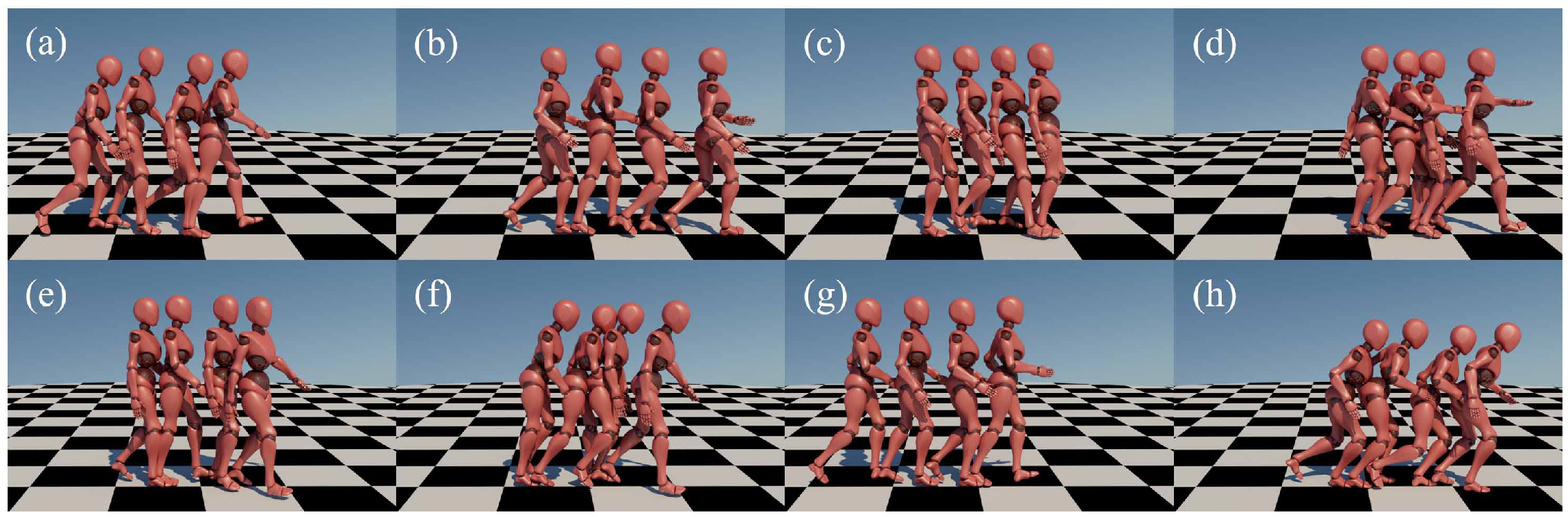, width = 1.0\linewidth}}
  \caption{Walking motion with styles. (a) is angry walking. (b) is sexy walking. (c) is proud walking. (d) is strutting walking. (e) is neutral walking. (f) is depressed walking. (g) is childlike walking. (h) is old walking.}
  \label{fig:walking}
 \end{figure*}

We compare our method with an existing DDPM-based method~\cite{findlay2022denoising}, and the DDPM baseline \cite{ho2020denoising} on the dataset proposed by \cite{xia2015realtime}. All three methods have been trained using the same hyperparameter setting as shown in Table \ref{tab:hyperparams}. The FID score is computed by the same amount of generated data with the same pre-trained classifier. The results have been reported in Table \ref{tab:quantitative}.
\begin{table}[htbp]
\caption{Quantitative comparison.}\label{tab:quantitative} \centering
\begin{tabular}{cc}
  \hline
  Model & FID ($\downarrow$)\\
  \hline
  \cite{findlay2022denoising} & 158.47\\
  \cite{ho2020denoising} & 198.67\\
  \hline
  Ours & \textbf{56.73}\\
  \hline
\end{tabular}
\end{table}
As shown in Table \ref{tab:quantitative}, our method achieves the best FID score compared with previous studies. This means the distribution of the generated data by our method has less difference from the ground truth dataset. In addition, we observe that the low performance in \cite{ho2020denoising} is mainly due to the lack of constraints on motion-specific aspects. Although \cite{findlay2022denoising} adds the adversarial training, it still does not perform well because it suffers from low motion quality in non-walking motion caused by its simple network design. Based on the observations above, our method achieves the best generative result with our proposed multi-task DDPM.

\subsection{Qualitative evaluations}

Apart from the quantitative comparison, we also provide several qualitative evaluations. After training is completed, we use the our model to generate motion samples. The number of total frames is kept the same as that of the training dataset, which is 32 frames for each motion clip. Figure \ref{fig:content} has shown the generated motions with different contents from our model. Our results show that our method is capable of generating different motions with different contents.

\begin{figure}[!h]
  \centering
   {\epsfig{file = 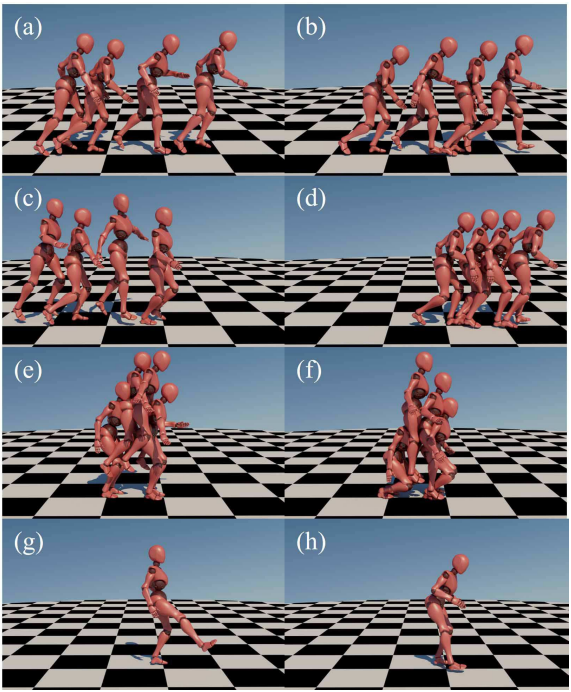, width = 1.0\linewidth}}
  \caption{We also provide generated styled motions with other contents. (a) is angry running. (b) is depressed running. (c) is strutting running. (d) is old running. (e) is sexy jumping. (f) is proud jumping. (g) is angry kicking. (h) is old kicking.}
  \label{fig:other}
 \end{figure}

In Figure \ref{fig:walking}, we have shown several sampled instances of styled walking motions. Our method produces different walking motions with styles, e.g. the old walking motion in Figure \ref{fig:walking}. As our method is based on stochastic generation, the capacity of our model is adequate for modelling various styles. Additionally, we also provide styled motions with other contents, such as running, in Figure \ref{fig:other}.

\subsection{Ablation Study}

We perform an ablation study to evaluate the effectiveness of multi-task architecture design. We use the same dataset proposed by \cite{xia2015realtime} and keep the same hyperparameter settings as shown in Table \ref{tab:hyperparams}. The FID score has been reported on the same amount of generated data and the same classifier.

\begin{table}[htbp]
\caption{Ablation study on multi-task architecture.}\label{tab:ablation}\centering
\begin{tabular}{cc}
  \hline
  Model & FID ($\downarrow$)\\
  \hline
  w/o foot loss & 74.68\\
  w/o root loss & 118.20\\
  w/o physical loss & 139.44\\
  w/o discriminator & 106.29\\
  \hline
  Ours - full & \textbf{56.73}\\
  \hline
\end{tabular}
\end{table}

As shown in Table \ref{tab:ablation}, our method achieves the best FID score among different ablations. From the results of the ablation study, we validate the effectiveness of our multi-task DDPM architecture. We observe a performance drop on the FID score by removing the discriminator. In our design, the discriminator harmonizes the separately estimated components for local guidance. The adversarial loss encourages to generate natural and coherent motions. If we remove the root prediction, the performance also drops a lot. As previously discussed, the trajectory of movements, which are represented by root, are mostly represented in the inter-class motion content. This result shows that human motions are heavily dependent on the trajectory of movements. The physical loss, including velocity and acceleration, provides another important motion information, which is represented by the intra-class styles. Within the same content, the difference between the actions mainly happens in the joint movement where velocity and acceleration can describe the intra-class motion styles. Our method without the foot loss still does not perform well. This is because estimating supporting foot patterns helps our neural network generate natural and coherent motions. Therefore, our ablation validates that each component in our multi-task DDPM architecture is effective towards generating high-quality motions.

\section{\uppercase{Conclusion}}
Styled motion synthesis is a critical and challenging problem with broad applications in many areas. We propose an end-to-end framework for styled motion synthesis where we successfully integrate two separate tasks, i.e. human motion synthesis and motion style transfer, into one pipeline. We represent our motions as the inter-class content component and the intra-class style component coupled in a common latent space. Such integration brings benefits more in optimal motion generation and thorough exploration of the coupled content-style latent space than previously two-staged frameworks. As motions contain a vast span of potential instances, we propose to leverage diffusion models in our framework to capture the diversity of motion data. We take DDPM for our implementation while our framework is generally compatible with other diffusion models. Our DDPM is designed to be a multi-task architecture for styled motion synthesis. The multi-task system is optimized by the local guidance including joint angles, trajectory of movements, supporting foot patterns and the global guidance including physical regulations and a discriminator based on the reconstruction formulation to encourage the generation of natural and coherent motions. Both qualitative and quantitative experimental results have shown the superior performance of our method. Our ablation study validates the effectiveness of our proposed multi-task architecture.

Our proposed pipeline is generally compatible with various diffusion-based generative models. In addition, our pipeline has the potential to model other motions (e.g., animal motions). However, as the motions become increasingly diverse, we may need new generative models with higher learning capacity to model the coupled content-style representations.

There are also a few potential future directions towards the styled human motion synthesis with diffusion models. One future work is that we will try our pipeline with a larger set of solutions containing also non-diffusion solutions and also try longer clip synthesis.

A future direction could be the stability and controllingness of the generation process via a stochastic system. Although the capacity of models has been greatly increased, the generation process is inherently trained under heavy uncertainty brought by noise injection. The stochastic property makes the generation difficult to achieve higher controllingness and stability. The stochastic system such as DDPM requires further mathematical theories on stochasticity to achieve higher stability and better controllingness.

\begin{figure}[!h]
  \centering
   {\epsfig{file = 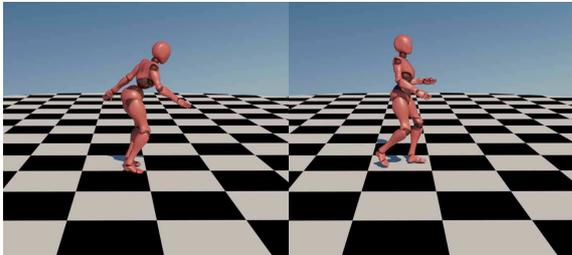, width = 1.0\linewidth}}
  \caption{Punching is a challenging case.}
  \label{fig:challenge}
 \end{figure}

Another potential direction in future is the control of limb ends, e.g. feet and hands. Figure \ref{fig:challenge} shows a challenging case. In human motions, the ends of limbs usually have tiny but complicated movements that DDPM is hard to model. Although we estimate foot patterns and trajectory of movements, the control of limb ends still has the potential to be improved by several ways such as physics-based guidance.

\bibliographystyle{apalike}
{\small
\bibliography{example}}

\begin{thebibliography}{}

\bibitem[Aberman et~al., 2020]{aberman2020unpaired}
Aberman, K., Weng, Y., Lischinski, D., Cohen-Or, D., and Chen, B. (2020).
\newblock Unpaired motion style transfer from video to animation.
\newblock {\em ACM Transactions on Graphics (TOG)}, 39(4):64--1.

\bibitem[Ahn et~al., 2018]{ahn2018text2action}
Ahn, H., Ha, T., Choi, Y., Yoo, H., and Oh, S. (2018).
\newblock Text2action: Generative adversarial synthesis from language to
  action.
\newblock In {\em 2018 IEEE International Conference on Robotics and Automation
  (ICRA)}, pages 5915--5920. IEEE.

\bibitem[Bond-Taylor et~al., 2021]{bond2021deep}
Bond-Taylor, S., Leach, A., Long, Y., and Willcocks, C.~G. (2021).
\newblock Deep generative modelling: A comparative review of vaes, gans,
  normalizing flows, energy-based and autoregressive models.
\newblock {\em arXiv preprint arXiv:2103.04922}.

\bibitem[Dhariwal and Nichol, 2021]{dhariwal2021diffusion}
Dhariwal, P. and Nichol, A. (2021).
\newblock Diffusion models beat gans on image synthesis.
\newblock {\em Advances in Neural Information Processing Systems},
  34:8780--8794.

\bibitem[Dong et~al., 2020]{dong2020adult2child}
Dong, Y., Aristidou, A., Shamir, A., Mahler, M., and Jain, E. (2020).
\newblock Adult2child: Motion style transfer using cyclegans.
\newblock In {\em Proceedings of the 13th ACM SIGGRAPH Conference on Motion,
  Interaction and Games}, MIG '20, New York, NY, USA. Association for Computing
  Machinery.

\bibitem[Findlay et~al., 2022]{findlay2022denoising}
Findlay, E.~J., Zhang, H., Chang, Z., and Shum, H.~P. (2022).
\newblock Denoising diffusion probabilistic models for styled walking
  synthesis.
\newblock {\em arXiv preprint arXiv:2209.14828}.

\bibitem[Henter et~al., 2020]{henter2020moglow}
Henter, G.~E., Alexanderson, S., and Beskow, J. (2020).
\newblock Moglow: Probabilistic and controllable motion synthesis using
  normalising flows.
\newblock {\em ACM Transactions on Graphics (TOG)}, 39(6):1--14.

\bibitem[Heusel et~al., 2017]{heusel2017gans}
Heusel, M., Ramsauer, H., Unterthiner, T., Nessler, B., and Hochreiter, S.
  (2017).
\newblock Gans trained by a two time-scale update rule converge to a local nash
  equilibrium.
\newblock {\em Advances in neural information processing systems}, 30.

\bibitem[Ho et~al., 2020]{ho2020denoising}
Ho, J., Jain, A., and Abbeel, P. (2020).
\newblock Denoising diffusion probabilistic models.
\newblock {\em Advances in Neural Information Processing Systems},
  33:6840--6851.

\bibitem[Holden et~al., 2017]{holden2017phase}
Holden, D., Komura, T., and Saito, J. (2017).
\newblock Phase-functioned neural networks for character control.
\newblock {\em ACM Transactions on Graphics (TOG)}, 36(4):1--13.

\bibitem[Kawar et~al., 2022]{kawar2022denoising}
Kawar, B., Elad, M., Ermon, S., and Song, J. (2022).
\newblock Denoising diffusion restoration models.
\newblock {\em arXiv preprint arXiv:2201.11793}.

\bibitem[Ling et~al., 2020]{ling2020character}
Ling, H.~Y., Zinno, F., Cheng, G., and Van De~Panne, M. (2020).
\newblock Character controllers using motion vaes.
\newblock {\em ACM Transactions on Graphics (TOG)}, 39(4):40--1.

\bibitem[Martinez et~al., 2017]{martinez2017human}
Martinez, J., Black, M.~J., and Romero, J. (2017).
\newblock On human motion prediction using recurrent neural networks.
\newblock In {\em Proceedings of the IEEE conference on computer vision and
  pattern recognition}, pages 2891--2900.

\bibitem[Mourot et~al., 2022]{mourot2022survey}
Mourot, L., Hoyet, L., Le~Clerc, F., Schnitzler, F., and Hellier, P. (2022).
\newblock A survey on deep learning for skeleton-based human animation.
\newblock In {\em Computer Graphics Forum}, volume~41, pages 122--157. Wiley
  Online Library.

\bibitem[Nichol et~al., 2021]{nichol2021glide}
Nichol, A., Dhariwal, P., Ramesh, A., Shyam, P., Mishkin, P., McGrew, B.,
  Sutskever, I., and Chen, M. (2021).
\newblock Glide: Towards photorealistic image generation and editing with
  text-guided diffusion models.
\newblock {\em arXiv preprint arXiv:2112.10741}.

\bibitem[Nichol and Dhariwal, 2021]{nichol2021improved}
Nichol, A.~Q. and Dhariwal, P. (2021).
\newblock Improved denoising diffusion probabilistic models.
\newblock In {\em International Conference on Machine Learning}, pages
  8162--8171. PMLR.

\bibitem[Petrovich et~al., 2021]{petrovich2021action}
Petrovich, M., Black, M.~J., and Varol, G. (2021).
\newblock Action-conditioned 3d human motion synthesis with transformer vae.
\newblock In {\em Proceedings of the IEEE/CVF International Conference on
  Computer Vision}, pages 10985--10995.

\bibitem[Ramesh et~al., 2022]{ramesh2022hierarchical}
Ramesh, A., Dhariwal, P., Nichol, A., Chu, C., and Chen, M. (2022).
\newblock Hierarchical text-conditional image generation with clip latents.
\newblock {\em arXiv preprint arXiv:2204.06125}.

\bibitem[Saharia et~al., 2022]{saharia2022photorealistic}
Saharia, C., Chan, W., Saxena, S., Li, L., Whang, J., Denton, E., Ghasemipour,
  S. K.~S., Ayan, B.~K., Mahdavi, S.~S., Lopes, R.~G., et~al. (2022).
\newblock Photorealistic text-to-image diffusion models with deep language
  understanding.
\newblock {\em arXiv preprint arXiv:2205.11487}.

\bibitem[Sohl-Dickstein et~al., 2015]{sohl2015deep}
Sohl-Dickstein, J., Weiss, E., Maheswaranathan, N., and Ganguli, S. (2015).
\newblock Deep unsupervised learning using nonequilibrium thermodynamics.
\newblock In {\em International Conference on Machine Learning}, pages
  2256--2265. PMLR.

\bibitem[Song et~al., 2020]{song2020denoising}
Song, J., Meng, C., and Ermon, S. (2020).
\newblock Denoising diffusion implicit models.
\newblock {\em arXiv preprint arXiv:2010.02502}.

\bibitem[Vahdat et~al., 2021]{vahdat2021score}
Vahdat, A., Kreis, K., and Kautz, J. (2021).
\newblock Score-based generative modeling in latent space.
\newblock {\em Advances in Neural Information Processing Systems},
  34:11287--11302.

\bibitem[Xia et~al., 2015]{xia2015realtime}
Xia, S., Wang, C., Chai, J., and Hodgins, J. (2015).
\newblock Realtime style transfer for unlabeled heterogeneous human motion.
\newblock {\em ACM Transactions on Graphics (TOG)}, 34(4):1--10.

\bibitem[Xiao et~al., 2021]{xiao2021tackling}
Xiao, Z., Kreis, K., and Vahdat, A. (2021).
\newblock Tackling the generative learning trilemma with denoising diffusion
  gans.
\newblock {\em arXiv preprint arXiv:2112.07804}.

\bibitem[Ye et~al., 2021]{ye2021human}
Ye, Z., Wu, H., and Jia, J. (2021).
\newblock Human motion modeling with deep learning: A survey.
\newblock {\em AI Open}.

\end{thebibliography}

\end{document}